\documentclass{article}

\usepackage{placeins}

\usepackage{subcaption}
\usepackage{arxiv}

\usepackage[utf8]{inputenc} 
\usepackage[T1]{fontenc}    
\usepackage{hyperref}       
\usepackage{url}            
\usepackage{booktabs}       
\usepackage{amsfonts}       
\usepackage{nicefrac}       
\usepackage{microtype}      
\usepackage{lipsum}		
\usepackage{graphicx}
\usepackage{doi}
\usepackage{CJKutf8}
\usepackage{multirow}

\usepackage{changepage}

\title{Cross-Corpus Multilingual Speech Emotion Recognition: Amharic vs. Other Languages}

\author{ \mbox{\href{https://orcid.org/0000-0003-0906-9605}{\includegraphics[scale=0.06]{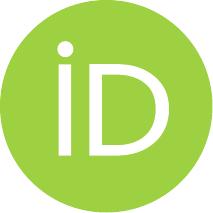}\strut}}\hspace{1mm}Ephrem Afele Retta \\
	School of Information Science and Technology\\
	Northwest University\\
	Xi’an 710127, China \\
	\texttt{afele@stumail.nwu.edu.cn} \\
	\And
	\mbox{\href{https://orcid.org/0000-0002-5549-5691}{\includegraphics[scale=0.06]{orcid.pdf}\strut}}\hspace{1mm}Richard Sutcliffe\thanks{Corresponding author} \\
	School of Computer Science and Electronic Engineering \\ University of Essex, Wivenhoe Park, Colchester CO4 3SQ, UK\\
	School of Information Science and Technology\\
	Northwest University, Xi’an 710127, China \\
	\texttt{rsutcl@essex.ac.uk, rsutcl@nwu.edu.cn} \\
	\And
	\mbox{\href{https://orcid.org/0000-0002-2872-0734}{\includegraphics[scale=0.06]{orcid.pdf}\strut}}\hspace{1mm}Jabar Mahmood \\
    Faculty of Computing and Information Technology \\
    University of Sialkot, Punjab, Pakistan \\
	School of Information and Engineering \\
	Chang'an University, Xi’an 710064, China \\
	\texttt{jabarmehmood@outlook.com, 2019024906@chd.edu.cn} \\
	\And
	\mbox{\href{https://orcid.org/0000-0002-5769-6040}{\includegraphics[scale=0.06]{orcid.pdf}\strut}}\hspace{1mm}Michael Abebe Berwo \\
	School of Information Science and Technology\\
	Chang'an University\\
	Xi’an 710064, China \\
	\texttt{2019024902@chd.edu.cn} \\
	\And
	\mbox{\href{https://orcid.org/0000-0002-7182-9639}{\includegraphics[scale=0.06]{orcid.pdf}\strut}}\hspace{1mm}Eiad Almekhlafi \\
	School of Information Science and Technology\\
	Northwest University\\
	Xi’an 710127, China \\
	\texttt{ealmekhlafi@stumail.nwu.edu.cn} \\
	\And
	\mbox{\href{https://orcid.org/0000-0002-0787-1357}{\includegraphics[scale=0.06]{orcid.pdf}\strut}}\hspace{1mm}Sajjad Ahmed Khan \\
	Computer Engineering Department \\
    Hoseo University \\
    Asan 31499, Republic of Korea \\
	\texttt{dr.sajjadkhan19@gmail.com} \\
	\And
	\mbox{\href{https://orcid.org/0000-0002-9321-6956}{\includegraphics[scale=0.06]{orcid.pdf}\strut}}\hspace{1mm}Shehzad Ashraf Chaudhry \\
    Department of Computer Science \& Information Technology \\
    College of Engineering, Abu Dhabi University \\
    Abu Dhabi, United Arab Emirates \\
    Department of Computer Engineering \\
	Nisantasi University \\
	Istanbul 34398, Turkey \\
	\texttt{ashraf.shehzad.ch@gmail.com} \\
	\And
	\mbox{\href{https://orcid.org/0000-0002-3106-669X}{\includegraphics[scale=0.06]{orcid.pdf}\strut}}\hspace{1mm}Mustafa Mhamed \\
	School of Information Science and Technology\\
	Northwest University\\
	Xi’an 710127, China \\
	College of Information and Electrical Engineering \\
	China Agricultural University \\
	Beijing 100083, China \\
	\texttt{mustafamhamed@stumail.nwu.edu.cn} \\
	\And
	\mbox{\href{https://orcid.org/0000-0002-0706-2103}{\includegraphics[scale=0.06]{orcid.pdf}\strut}}\hspace{1mm}Jun Feng\thanks{Corresponding author}\\
	School of Information Science and Technology \\
	Northwest University\\
	Xi’an 710127, China \\
	\texttt{fengjun@nwu.edu.cn} \\
}
\date{}

\renewcommand{\headeright}{E. A. Retta et al.}
\renewcommand{\undertitle}{}
\renewcommand{\shorttitle}{Cross-Corpus Multilingual Speech Emotion Recognition}

\hypersetup{
pdftitle={Cross-Corpus Multilingual Speech Emotion Recognition: Amharic vs. Other Languages},
pdfsubject={cs.SD, eess.AS, cs.LG},
pdfauthor={Ephrem Afele Retta, Richard Sutcliffe, Jabar Mahmood, Michael Abebe Berwo, Eiad Almekhlafi, Sajjad Ahmed Khan, Shehzad Ashraf Chaudhry, Mustafa Mhamed, Jun Feng},
pdfkeywords={Speech emotion recognition, Convolutional neural network, cross-corpus, multilingual, multiple training languages, Amharic},
}

\begin{document}
\maketitle

\begin{abstract}
In a conventional Speech emotion recognition (SER) task, a classifier for a given language is trained on a pre-existing dataset for that same language. However, where training data for a language does not exist, data from other languages can be used instead.
We experiment with cross-lingual and multilingual SER, working with Amharic, English, German and URDU. For Amharic, we use our own publicly-available Amharic Speech Emotion Dataset (ASED). For English, German and Urdu we use the existing RAVDESS, EMO-DB and URDU datasets. We followed previous research in mapping labels for all datasets to just two classes, positive and negative. Thus we can compare performance on different languages directly, and combine languages for training and testing. In Experiment 1, monolingual SER trials were carried out using three classifiers, AlexNet, VGGE (a proposed variant of VGG), and ResNet50. Results averaged for the three models were very similar for ASED and RAVDESS, suggesting that Amharic and English SER are equally difficult. Similarly, German SER is more difficult, and Urdu SER is easier.
In Experiment 2, we trained on one language and tested on another, in both directions for each pair: Amharic$\leftrightarrow$German, Amharic$\leftrightarrow$English, and Amharic$\leftrightarrow$Urdu. Results with Amharic as target suggested that using English or German as source will give the best result.
In Experiment 3, we trained on several non-Amharic languages and then tested on Amharic. The best accuracy obtained was several percent greater than the best accuracy in Experiment 2, suggesting that a better result can be obtained when using two or three non-Amharic languages for training than when using just one non-Amharic language.
Overall, the results suggest that cross-lingual and multilingual training can be an effective strategy for training a SER classifier when resources for a language are scarce.
\end{abstract}

\keywords{Speech emotion recognition \and Convolutional neural network \and cross-corpus \and multilingual \and multiple training languages \and Amharic}

\section{Introduction}
Emotions assist individuals to communicate and to comprehend others' points of view \cite{zvarevashe2020ensemble}.
Speech emotion recognition (SER) is the task of comprehending emotion in a voice signal, regardless of its semantic content \cite{khan2020novel}. 

SER datasets are not available in all languages. Moreover, the quantity and quality of the training data which is available varies considerably from one language to another. For example,  when evaluated across several datasets, differences in corpus language, speaker age, labeling techniques, and recording settings significantly influence model performance \cite{zhang2016cross, zhang2011unsupervised}. This encourages the development of more robust SER systems capable of identifying emotion from data in different languages. This can then permit the implementation of voice-based emotion recognition systems in real-time for an extensive variety of industrial and medical applications.

The majority of research on SER has concentrated on a single corpus, without considering cross-lingual and cross-corpus effects. One reason is that, in comparison to the list of spoken languages, we only have a small number of corpora for the study of speech analysis \cite{wang2015transfer}.
Furthermore, even when only considering the English language, accessible resources vary in quality and size, resulting in the dataset sparsity problem observed in SER research. In such instances, learning from a single data source makes it challenging for SER to function effectively. As a result, more adaptable models that can learn from a wide range of resources in several languages are necessary for practical applications.

Several researchers have investigated cross-corpus SER in order to enhance classification accuracy across several languages. These works employed a variety of publicly accessible databases to highlight the most interesting trends \cite{schuller2010cross}.
Even though some research has addressed the difficulty of cross-corpus SER, as described in  Schuller et al. \cite{schuller2010cross}, the challenges posed by minority languages such as Amharic have not been investigated.

Amharic is the second-largest Semitic language in the world after Arabic and it also the national language of Ethiopia \cite{mossie2018social}. In terms of the number of speakers and the significance of its politics, history, and culture, it is one of the 55 most important languages in the world  \cite{mengistu2017text}. Dealing with such languages is critical to the practicality of next-generation systems \cite{albornoz2015emotion}, which must be available for many languages.

In previous work \cite{retta2023new}, we created tbe first spontaneous emotional dataset for Amharic. This contains 2,474 recordings made by 65 speakers (25 male, 40 female) and uses five emotions: fear, neutral,  happy, sad, and angry. The Amharic Speech Emotion Dataset (ASED) is publicly available for download\footnote{\url{ https://github.com/Ethio2021/ASED_V1}}. 
The ASED dataset allows us to carry out the work reported here.

The contributions of this paper are as follows:
\begin{itemize}
\item We investigate different scenarios for monolingual, cross-lingual and multilingual SER using datasets for Amharic and three other languages (English, German and Urdu).


\item We experiment with a novel approach in which a model is trained on data in several non-Amharic languages before being tested on Amharic. We show that training on two non-Amharic languages gives a better result than training on just one.

\item We present a comparison of deep learning techniques in these tasks: AlexNet, ResNet50, and VGGE.

\item This is the first work that shows the performance tendencies of Amharic SER utilizing several languages, to the best of our knowledge.
\end{itemize}

The structure of this paper is as follows: Section 2 presents previous work. Section 3 explains our approach, datasets, and feature extraction methods for SER. Section 4 presents the proposed deep learning architecture and experimental settings. Section 5 describes the experiments and outcomes. Finally, Section 6 gives conclusions and next steps.

\begin{table}[t!]
\begin{adjustwidth}{0in}{-0.35in} 
\centering
\caption{Previous work on cross-lingual and multilingual SER (X=Cross-lingual, M=Multilingual, SVM=Support Vector Machine, SC=Standard Classifier, EPC=Emotional Profile Classifier, SMO=Sequential Minimal Optimization, DBN=Deep Belief Network, MLP=Multi-Layer Perceptron, GAN=Generative Adversarial Network, LSTM=Long Short-Term Memory, LR=Logistic Regression, RF=Random Forest, J48=Decision Tree).}
\begin{tabular}{|c|c|c|c|c|c|}
\hline
\bf{Ref} & 
\begin{tabular}[c]{@{}l@{}}\bf{Methods}\\ \bf{Employed}\end{tabular} &
\begin{tabular}[c]{@{}l@{}}\bf{Feature} \\ \bf{Extraction}\end{tabular} & 
\begin{tabular}[c]{@{}l@{}}\bf{Databases \&} \\\bf{Languages} \end{tabular} & 
\begin{tabular}[c]{@{}l@{}}\bf{Expts}\end{tabular} &
\begin{tabular}[c]{@{}l@{}}\bf{Classes}\end{tabular}
\\ \hline

 
\cite{lefter2010emotion} & 
 SVM & 
 \begin{tabular}[c]{@{}c@{}}Prosodic\end{tabular} & 
 \begin{tabular}[c]{@{}c@{}}EMO-DB (German), DES (Danish)\\ENT (English), SA (Afrikaans)\end{tabular} &
 XM &
 \begin{tabular}[c]{@{}l@{}}3 \end{tabular}
\\

\cite{albornoz2015emotion} & 
 SVM, SC, EPC & 
 \begin{tabular}[c]{@{}c@{}}Prosodic\\Various\\MFCC\end{tabular} & 
 \begin{tabular}[c]{@{}c@{}}RML (Mandarin, English, Italian\\Persian, Punjabi, Urdu)\end{tabular} &
 X &
 \begin{tabular}[c]{@{}l@{}}6 \end{tabular}
\\

\cite{ xiao2016speech} & 
 SMO & 
 \begin{tabular}[c]{@{}c@{}}Various\\MFCC\end{tabular} & 
 \begin{tabular}[c]{@{}c@{}}CDESD (Mandarin), EMO-DB, DES\end{tabular} &
 X &
 \begin{tabular}[c]{@{}c@{}}Arousal\\Appraisal\\Space \end{tabular}
\\

\cite{sagha2016enhancing} & 
 SVM & 
 \begin{tabular}[c]{@{}c@{}}Various\\MelSpec\end{tabular} & 
 \begin{tabular}[c]{@{}c@{}}EU-EmoSS (English, French, German,\\ Spanish), VESD (Chinese), CASIA (Chinese)\end{tabular} &
 X &
 \begin{tabular}[c]{@{}c@{}}Arousal\\Valence\\Plane \end{tabular}
\\

\cite{meftah2017cross} & 
 DBN, MLP & 
 \begin{tabular}[c]{@{}c@{}}Low-level\\ Acoustic\end{tabular} & 
 \begin{tabular}[c]{@{}c@{}}KSUEmotions (Arabic), EPST (English)\end{tabular} &
 X &
 \begin{tabular}[c]{@{}l@{}}2 \end{tabular}
 \\
 
\cite{latif2018cross} & 
SVM &
eGeMAPS & 
\begin{tabular}[c]{@{}c@{}}SAVEE (English), EMOVO (Italian),\\EMO-DB, URDU (Urdu)\end{tabular} &
XM &
\begin{tabular}[c]{@{}l@{}}2 \end{tabular}
\\

\cite{latif2018cross1} & 
DBN & 
eGeMAPS & 
\begin{tabular}[c]{@{}c@{}}FAU-AIBO (German), IEMOCAP (English)\\EMO-DB, SAVEE, EMOVO\end{tabular} &
X &
\begin{tabular}[c]{@{}l@{}}2 \end{tabular}
\\

\cite{latif2019unsupervised} & 
 GAN & 
 \begin{tabular}[c]{@{}c@{}}eGeMAPS\\Various\end{tabular} & 
 \begin{tabular}[c]{@{}c@{}}EMO-DB, SAVEE, EMOVO, URDU\end{tabular} &
 XM &
 \begin{tabular}[c]{@{}l@{}}2 \end{tabular}
\\

\cite{goel2020cross} & 
 LSTM, LR, SVM & 
 \begin{tabular}[c]{@{}c@{}}ISO9\end{tabular} & 
 \begin{tabular}[c]{@{}c@{}}EMOVO, EMO-DB, SAVEE, IEMOCAP,\\ MASC (Chinese)\end{tabular} &
 X &
 \begin{tabular}[c]{@{}l@{}} 5 \end{tabular} 
\\
\cite{duret2023learning} &   \begin{tabular}[c]{@{}l@{}}CNN and \\ Wav2Vec2-XLSR\end{tabular} & prosody &  \begin{tabular}[c]{@{}c@{}}IEMOCAP, CREMA-D (English), ESD (English), \\ Synpaflex (French), Oreau (French), \\EMO-DB, EMOVO, emoUERJ (Portuguese)\end{tabular} & X & 4
\\
\cite{pandey2023multi} & AG-TFNN & MelSpec & \begin{tabular}[c]{@{}c@{}}EMO-DB, eNTERFACE (English), \\ IITKGP-SEHSC (Hindi), IITKGP-SESC (Telugu), \\ ShEMO-DB (Persian)\end{tabular} & M & 2

\\
\cite{zehra2021cross} & 
\begin{tabular}[c]{@{}l@{}}SMO, RF, J48 \\Ensemble\end{tabular} & 
\begin{tabular}[c]{@{}c@{}}Spectral \\Prosodic\\ eGeMAPS\end{tabular} & 
\begin{tabular}[c]{@{}c@{}} SAVEE, URDU, EMO-DB, EMOVO\end{tabular} &
X &
\begin{tabular}[c]{@{}l@{}}2 \end{tabular}
\\\hline
\end{tabular}
\label{previous-work}
\end{adjustwidth}
\end{table}

\section{Related Work}
Over the last two decades, much important research has been conducted on speaker-independent SER. This work has shown that several factors influence accuracy, including the dataset utilized, the features extracted, and the classifier network employed to predict emotions. Sailunaz el al. \cite{sailunaz2018emotion} present a thorough survey of the datasets available, the features extracted, and the networks most commonly employed for SER. However, while there has been preliminary research on enhancing the robustness of SER by combining multiple emotional speech corpora to form the training set and thereby minimising data scarcity, there is a shortage of studies on multilingual cross-corpus SER \cite{schuller2010cross,schuller2011recognising}. In the following, we first summarize related cross-lingual work. After that we outline multilingual studies. Information about all the research is shown in Table \ref{previous-work}. 

Concerning cross-lingual studies,
Lefter et al. \cite{lefter2010emotion} carried out an early study in which they trained a SER classifier on one or more datasets and then tested on another. In a cross-lingual setting, training on ENT and testing on DES gave the lowest Equal Error Rate for Anger (29.9\%).



Albornoz et al. \cite{albornoz2015emotion} proposed a SER classifier for emotion detection, focusing on emotion identification in unknown languages. The results showed what could be expected from a system trained with a different language, reaching 45\% on average. The standard multi-class SVM performed better than the classifier implemented using Emotion Profiles (EP). On average, the Standard Classifier (SC) reached 56.8\%, whereas the Emotional Profile Classifier (EPC) obtained 52.1\%.

Xiao et al. \cite{xiao2016speech} examined SER for Mandarin Chinese vs. Western languages like German and Danish. The authors concentrated on gender-specific SER and attained classification rates that were higher than chance but lower than baseline accuracy. The best classification rate in the cross-language family test on male speech samples (71.62\%), was when the Chinese Dual mode Emotional Speech Database (CDESD) was used for training and Emo-DB was used for testing.

Sagha et al. \cite{sagha2016enhancing} utilized language detection to improve cross-lingual SER. They found that using a language identifier followed by network selection rather than a network trained on all existing languages was superior for recognizing the emotions of a speaker whose language is unknown. On average, the Language IDentification (LID) approach for selecting training corpora was superior to using all the available corpora when the spoken language was not known.

Meftah et al. \cite{meftah2017cross} proposed Deep Belief Networks (DBN) for cross-corpus SER and evaluated them in comparison with MLP via emotional speech corpora for Arabic (KSUEmotions) and English (EPST). Training on one dataset and testing on the other yielded similar results for both directions and both models. The best result was Arabic$\rightarrow$English using DBN (Valence 53.22\%, Arousal 57.2\%).

Latif et al. \cite{latif2018cross} extracted eGeMAPS features from their raw audio data. They used SVM with a Gaussian kernel for classifying data into their respective categories. The best result was when training on EMO-DB and then testing on URDU (57.87\%).

Latif et al. \cite{latif2018cross1} also used eGeMAPS features and they employed five different corpora for three different languages to investigate cross-corpus and cross-language emotion recognition using Deep Belief Networks (DBNs). IEMOCAP performs well on EMO-DB compared to FAU-AIBO even though both the latter datasets are German.


Latif et al. \cite{latif2019unsupervised} studied SER using languages from various language families, such as Urdu vs. Italian or German. The best cross-lingual results were obtained by training on URDU and testing on EMODB (65.3\%) and the worse were by training on URDU and testing on SAVEE (53.2\%).

Goel et al. \cite{goel2020cross} used transfer learning to carry out multi-task learning experiments and discovered that traditional machine learning architectures \cite{wang2015transfer, bhaykar2013speaker} can perform as well as deep learning neural networks for SER provided the researchers pick appropriate input features. Training the model on IEMOCAP and testing on EMO-DB obtained the best performance (65\%).

Zehra et al. \cite{zehra2021cross} presented an ensemble learning approach for cross-corpus machine learning SER, utilizing the SAVEE, URDU, EMO-DB, and EMOVO databases. The method employed three of the most prominent machine learning algorithms, Sequential Minimal Optimization (SMO), Random Forest (RF), and Decision Tree (J48), plus a majority voting mechanism. The ensemble approach was worse than the other classifiers except when training on EMOVO and testing on URDU (62.5\%).

Jarod et al. \cite{duret2023learning} used prosody prediction and
employed eight different corpora for five European
languages to investigate cross-lingual and multilingual emotion recognition using Wav2Vec2XLSR. The multilingual setup outperformed the monolingual one for all selected European languages, except English, by a very small margin.

Pandey et al. \cite{pandey2023multi} proposed a SER classifier for emotion detection, focusing on learning emotions, irrespective of culture. They also used 3D Mel-Spectrogram features (henceforth MelSpec) and employed five different corpora for five languages to investigate cross-lingual emotion recognition using an Attention-Gated Tensor Factorized Neural Network (AG-TFNN). The best result was Fold2$\rightarrow$German using 3D TFNN. In addition, Fold5$\rightarrow$Telugu had better performance when compared to Fold4$\rightarrow$Hindi, even though both languages are of Indian origin.

We now consider multilingual approaches in which several datasets in different languages are used for training. 
In addition to the cross-lingual experiments referred to earlier, Lefter et al. \cite{lefter2010emotion} also carried out some multilingual work in which they trained on various pairs or triples of datasets chosen from EMO-DB, DES and ENT, and tested on each of these individually. The best result was obtained by training on all three and testing on EMO-DB (Equal Error Rate 20.5\%).

Latif et. al \cite{latif2018cross} used four different corpora (SAVEE, EMOVO, EMO-DB and URDU) for four different languages to investigate multilingual emotion recognition using Support Vector Machines (SVM). When training on EMO-DB, EMOVO and SAVEE and testing on URDU, a result of 70.98\% was achieved, which was higher then any pair of these datasets.


Latif et al. \cite{latif2019unsupervised} also used SAVEE, EMOVO, EMO-DB and URDU. The best performance was training on SAVEE, EMOVO, URDU and testing on EMO-DB (68\%). The worst performance was training on the same three datasets and testing on EMOVO (61.8\%)

Regarding the model used, Latif et al. \cite{latif2018cross}, Albornoz et al. \cite{albornoz2015emotion},  Lefter et al. \cite{lefter2010emotion}, and Sagha et al. \cite{sagha2016enhancing} are all based on SVMs. Meftah et al. \cite{meftah2017cross} and Latif et al. \cite{latif2018cross1} utilized DBN, Goel et al. \cite{goel2020cross}, Jarod et al. \cite{duret2023learning} and Pandey et al. \cite{pandey2023multi} applied machine learning and deep learning methods, Zehra et al. \cite{zehra2021cross} used ensemble methods, and lastly Xiao et al. \cite{ xiao2016speech} and Latif et al. \cite{latif2019unsupervised} applied GAN, and SMO respectively.
Concerning the earlier studies we observe that the SVM algorithm performs poorly on large data sets. It also performs poorly in situations with more characteristics per data point, especially in multi-class situations. When attempting to extract features from DBN plus low-level acoustic information, vs. DBN with eGeMAPS, the latter significantly outperformed. Additionally, deep learning models outperform conventional classifiers. However, the model of Goel et al. \cite{goel2020cross} extracts features quite well but requires a lot of training time. As previously indicated, an ensemble strategy only provided the best performance in one scenario. Furthermore, the existing techniques in SER lack preprocessing operations. We conclude that, across many datasets, binary performance outperforms multiple classes.  Plus none of the previous work has focused on the Amharic language.

Here, we first present the preprocessing strategy before the extraction of features from the signal.
Second, we propose an architecture, based on the VGG model, which offers good results. 
Third, we provide a classification benchmark for Amharic and three non-Amharic languages using deep Neural Networks. Finally, we contrast the effectiveness of our novel training scenarios to demonstrate the efficiency of cross-lingual and multilingual approaches.

\section{Approach}
\begin{figure}
\begin{adjustwidth}{0in}{0in} 
			\centering
			\includegraphics[width=133mm,scale=2]{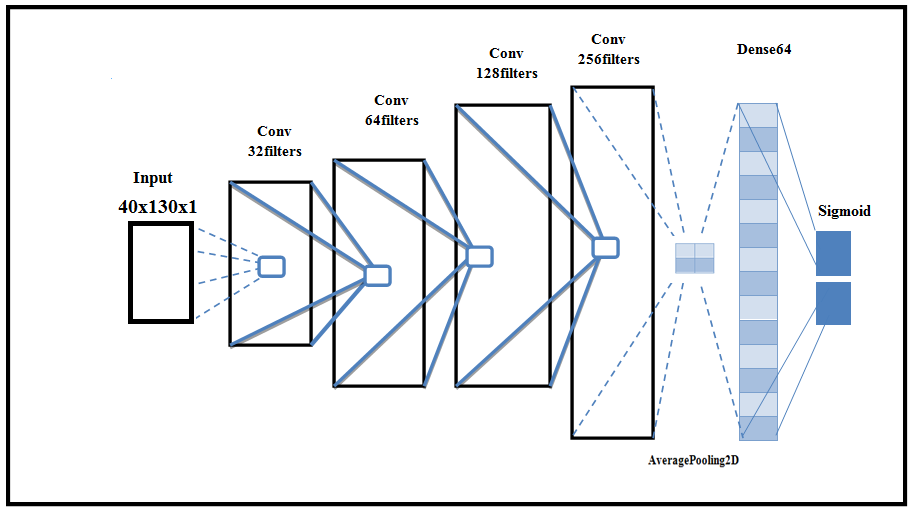}
			\caption{ Network architecture of proposed VGGE, based on well-known VGG model.}
			\label{block}
			\end{adjustwidth}
		\end{figure}

Many factors influence SER accuracy in a cross-corpus and multilingual context. The dataset utilized, the features extracted from the speech signals, and the neural network classifiers implemented to identify emotion are all essential aspects that might significantly impact the results.
Our SER method is summarized in Figure \ref{block}. We use four corpora (ASED, RAVDESS, EMO-DB, URDU) to test SER in Amharic, English, German and Urdu respectively.

One difficulty faced with this research is that datasets use different sets of emotion labels, as can be seen in Table \ref{comparison-ASED-RAVDESS-EMO-DB-URDU}. Following previous work \cite{deng2013sparse, eyben2015geneva} we address this by mapping labels onto just two classes, positive valence and negative valence, as indicated in the table.


Further details on the chosen datasets, feature extraction, and classifier are provided below.

\begin{figure}
\begin{adjustwidth}{0in}{0in} 
			\centering
			\includegraphics[width=133mm,scale=2]{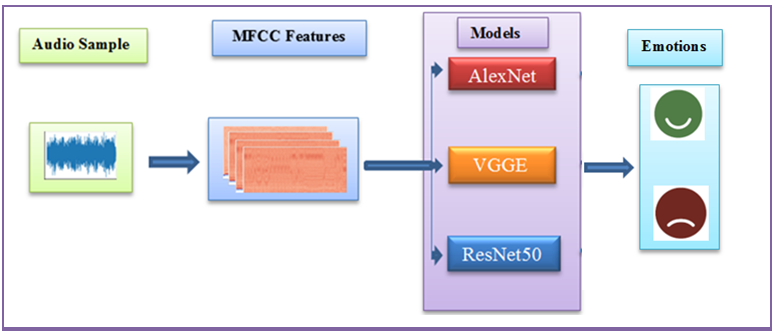}
			\caption{ Block diagram of our approach for SER.}
			\label{vgg-style-network}
			\end{adjustwidth}
		\end{figure}

\begin{table}[t!]
\begin{adjustwidth}{0in}{0in} 
\centering
\caption{
{\bf Datasets used in the experiments. The table also shows the mapping from the emotion labels in each dataset onto just two valence labels which can be used across them all: Positive and Negative.}}
\begin{tabular}{|c|c|c|c|c|}
\hline
{\bf Aspect} & {\bf ASED} & {\bf RAVDESS} & {\bf EMO-DB} & {\bf URDU} \\ \hline
Language  &Amharic & English        & German             & Urdu  \\ \hline
Recordings &2474  & 1440      & 535                   &  400\\ \hline
Sentences &27    &  2        & 10                      &  - \\ \hline
Participants & 65   &   24    & 10                     & 38\\ \hline
Emotions &5     &   8     & 7                      &  4\\ \hline
Positive valence & Neutral, Happy & \begin{tabular}[c]{@{}l@{}}{\small Neutral, Happy,}\\{\small Calm, Surprise}\end{tabular}   & Neutral, Happiness   &  Neutral, Happy\\ \hline
Negative valence & \begin{tabular}[c]{@{}l@{}}{\small Fear, Sadness,}\\{\small Angry}\end{tabular} &    \begin{tabular}[c]{@{}l@{}}{\small Fear, Sadness,}\\{\small  Angry, Disgust}\end{tabular}&\begin{tabular}[c]{@{}l@{}}{\small Anger, Sadness,}\\ {\small  Fear, Disgust,} \\{\small Boredom}\end{tabular}      & Angry, Sad\\ \hline
References & \cite{retta2023new} & \cite{livingstone2018ryerson} &      \cite{burkhardt2005database}  &  \cite{latif2018cross}\\ \hline
\end{tabular}
\label{comparison-ASED-RAVDESS-EMO-DB-URDU}
\end{adjustwidth}
\end{table}

\begin{table}[t!]
\begin{adjustwidth}{0in}{0in} 
\centering
\caption{
{\bf Statistics of clip lengths for all datasets (1-2.0 i.e. 1$<$=d$<$2).}}
\begin{tabular}{|c|c|c|c|c|}
\hline
{\bf Duration} & {\bf ASED} & {\bf EMO-DB} & {\bf RAVDESS} & {\bf URDU} \\ \hline
1-2.0     &     & 126  &     &   \\ \hline
2-3.0     & 850 & 224  &     & 200\\ \hline
3.0-4     & 1624& 136  & 1440& 200 \\ \hline
4.0-5     &     & 24   &     &  \\ \hline
5.0-6     &     & 20   &     & \\ \hline
6.0-7     &     & 3    &     &  \\ \hline
7.0-8     &     & 1    &     & \\ \hline
8.0-9     &     & 1    &     &\\ \hline
\textbf{STD}       & 0.444   & 1.067    & 0   & 0.5\\ \hline
\textbf{Mean}      & 2.967   & 2.267    & 3   &  2.5\\ \hline
\end{tabular}
\label{clip-lengths}
\end{adjustwidth}
\end{table}

\subsection{Speech emotion databases}\label{databases}	

\textbf{EMO-DB} \cite{burkhardt2005database} is for German, uses five emotions and contains ten everyday sentences, five made of one phrase, five made of two phrases. There are ten speakers (5 male, 5 female), nine qualified in acting and about 535 raw utterances in total. Recording is done at 16 kHz and 16 bits, and was carried out in the Anechoic chamber of the Technical Acoustics Department at the Technical University Berlin.

\textbf{RAVDESS} \cite{livingstone2018ryerson} is for English, uses eight emotions and contains just two sentences. The 24 speakers (12 male, 12 female) are professional actors. Interestingly, emotion in this dataset is `self-induced' \cite{stanislavski1936actor}, rather than Acted. Moreover, there are two levels of each emotion. There are 4,320 utterances.  Project investigators selected the best two clips for each speaker and each emotion. Recording was at 48 kHz and 16 bits, and it was carried out in a professional recording studio at Ryerson University.

\textbf{URDU} \cite{latif2018cross} is for Urdu, uses four emotions and comprises 400 audio recordings from Urdu TV talk shows. There are 38 speakers (27 male, 11 female). Emotions are not acted, but occur naturally during the conversations between guests on the talk shows.

\textbf{ASED} \cite{retta2023new}
is for Amharic and was created by the authors in previous work. It uses five emotions and consists of 2,474 recordings made by 65 speakers (25 male, 40 female). Recording is done at 16 kHz and 16 bits. The ASED dataset is accessible to the public for research purposes (see URL in earlier footnote).
\subsection{Data Preprocessing}

\begin{figure}
\begin{adjustwidth}{0in}{0in} 
			\centering
			\includegraphics[width=133mm,scale=2]{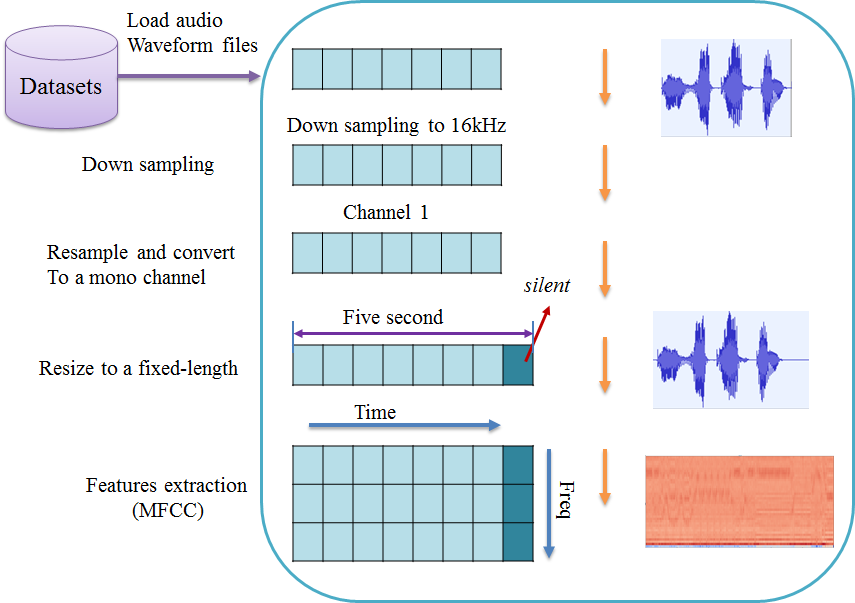}
			\caption{ Data Preprocessing.}
			\label{Preprocessing}
			\end{adjustwidth}
		\end{figure}

Before proceeding to feature extraction, a number of pre-processing steps were performed on the datasets as shown in Figure \ref{Preprocessing}. Recordings were first downsampled to 16kHz and converted to mono. Most of the sound clips in the datasets are 5 seconds in length or less. A few are longer than this. Therefore, we extended any shorter clips to 5 seconds by adding silence to the end. Conversely any longer clips were cut off in order to make them exactly 5 seconds long. The statistics of clip lengths are shown in Table \ref{clip-lengths}

\begin{figure}
\begin{adjustwidth}{0in}{0in} 
			\centering
			\includegraphics[width=133mm,scale=2]{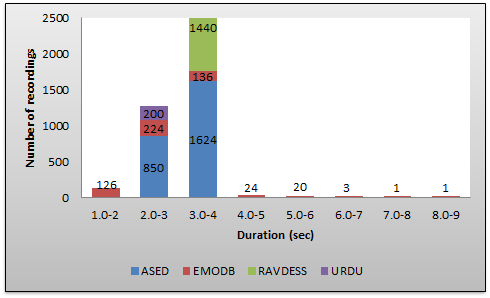}
			\caption{ Distribution of utterance lengths for all datasets, based on duration ranges. $2.0-3$ in the figure means $2s <= d < 3s$ where $d$ is the utterance length, and the same for the other ranges.}
			\label{duration_of_utterances}
			\end{adjustwidth}
		\end{figure}
\subsection{Feature extraction for SER}
A vast amount of information reflecting emotional characteristics is present in the speech signal. One of the key issues within SER research is the choice of features which should be used.

Previously, traditional feature extraction methods such as prosodic features were used for SER \cite{fairbanks1941experimental, gangamohan2016analysis}, including the variance, intensity, spoken word rate, and pitch. However, some traditional features are shared across different emotions, as discussed by Gangamohan et al. \cite{gangamohan2016analysis}.
For example, as observed in Table 11.2 of Gangamohan et al., angry and happy utterances have similar trends in F0 and speaking rate, compared to neutral speech.

Manually extracted traditional features may work well with traditional classification methods in machine learning, where a set of features or attributes describes each instance in a dataset \cite{khalil2019speech}. In contrast, however, deep learning can itself determine which features to focus on to recognize verbal emotions. Finding some set of feature vectors or properties that can give a compact representation of the input audio signal has therefore become the main aim of feature extraction methods. The spectrum extraction methods convert the input sound waveform to some discrete shape or feature vector. Normally, the speech signal is not static but when looking at a short period of time, it acts as a static signal. This short, detached snap is called a frame. The acoustic model extracts features from the frames \cite{dey2019acoustic, almekhlafi2022classification}. Feature extraction deals with obtaining useful information for reference by removing irrelevant information. These extracted feature vectors are fed into deep learning models. In short, spectrum extraction methods can convert audio signals into vectors that deep learning models can handle. The model can then be trained to learn the features of each emotion and hence classify it. Overall, this is one reason why deep learning models can perform better than machine learning models.

After reviewing many works on SER, it is clear that Mel-Frequency Cepstral Coefficients (MFCC) are widely used in audio classification and speech emotion recognition \cite{issa2020speech}.
 MFCC is a coefficient that expresses the short-term power spectrum of a sound. It uses a series of steps to imitate the human cochlea, thereby converting audio signals. The Mel scale is significant because it approximates the human perception of sound instead of being a linear scale \cite{shaw2016emotion}. In previous work \cite{retta2023new} we compared MFCC to alternatives and found it to be the best. This is the reason we choose MFCC features for the present study.
     
\section{Architectures and Settings} \label{architectures}
Most prior research uses CNN-based models for SER \cite{kwon2020cnn}. Among such models, the notable ones include AlexNet \cite{kumbhar2019speech}, VGG \cite{simonyan2014very,molchanov2016pruning}, and ResNet50 \cite{he2016deep,george2017deep}.
This section provides a short overview of the models. Our proposed model, VGGE, is a variant of VGG.

	\begin{itemize}	
       \item \textbf{AlexNet}
       is one of the famous CNN models used in applications such as image classification and recognition, and is widely employed for SER classification \cite{sajjad2020clustering}. It achieved an outstanding result at the ImageNet competition in 2012 \cite{krizhevsky2012imagenet}.
      \item\textbf{VGG} \cite{simonyan2014very} appeared in 2014, created by the Oxford Robotics Institute. It is well known that the early CNN layers capture the general features of sounds such as wavelength, amplitude, etc., and later layers capture more specific features such as the spectrum and the cepstral coefficients of waves. This makes a VGG-style model suitable for the SER task. After some experimentation, we found that a model based on VGG but using four layers gave the best performance. We call this proposed model VGGE and used it for our experiments. Figure \ref{block} shows the settings for VGGE. 
      \item \textbf{ResNet} \cite{he2016deep} was launched in late 2015. This was the first time that networks having more than a hundred layers were trained. Subsequently it has been applied to SER classification \cite{george2017deep}.
	\end{itemize}

Concerning the experimental setup, the standard code for AlexNet and ResNet50 was downloaded and used for the experiments. For VGGE, the network configuration was altered, as shown in Figure \ref{block}. For the other models, the standard network configuration and parameters were used.

In all experiments, the librosa v0.7.2 library \cite{sharmin2020bengali} was used to extract MFCC features.

We used the Keras deep learning library, version 2.0, with a Tensorflow 1.6.0 backend to build the classification models. The models were trained using a machine with an NVIDIA GeForce GTX 1050. Our model employed the Adam optimization algorithm, with categorical cross-entropy as the loss function; training was terminated after 100 epochs and batch size was set to 32.

\section{Experiments}

Mostly, two methods are utilized for speaker-independent SER \cite{shinde2021speech}: The first method is Leave One Speaker Out (LOSO) \cite{bhaykar2013speaker,deb2018multiscale,wang2020wavelet}; Here, when the corpus contains $n$ speakers, we use $n-1$ speakers for training and the remaining speaker for testing. For cross-validation, the experiment is repeated $n$ times with a different test speaker each time. In the second method, the training and testing sets have been determined previously \cite{latif2018cross, swain2015study, kuchibhotla2016optimal}.

In our work, we followed the second approach. For the first monolingual experiment (train on a corpus, test on the same corpus) the data was split into training,  testing, and validation sets randomly five times, each time ensuring that the split sets are speaker-independent.
As shown in Table \ref{distribution}, all the datasets were split 70\% train, 20\% test, and 10\% validation.
In the first experiment, we also carried out a sentence-independent study where sentences used for training were not used for testing.

The second and third experiments are the cross-lingual experiment (train on a corpus in one language, test on a corpus in another language) and the multilingual experiment (train on two or three corpora joined together, each in a different non-Amharic language, and test on the Amharic ASED corpus.
In these experiments, the speakers in the validation sets are not seen in the training sets. Moreover, the speakers in the testing set are by definition not the same as those in the training and validation sets, as they are from different datasets. 

\begin{figure}
\begin{adjustwidth}{0in}{0in} 
			\centering
			\includegraphics[width=133mm,scale=2]{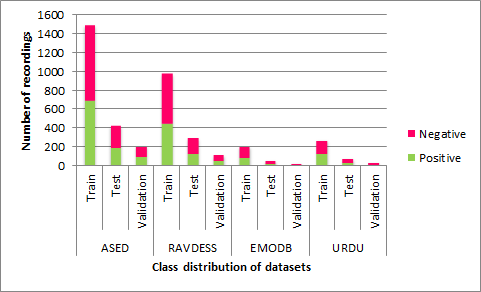}
			\caption{ Class distribution within Datasets.}
			\label{Class distribution}
			\end{adjustwidth}
		\end{figure}

Figure \ref{Class distribution} shows a label distribution that is balanced across partitions. 
The performance of the proposed classification of Amharic language data used in monolingual, cross-lingual, and multi-lingual SER experiments is evaluated using F1-score and accuracy. 
We have shared the file names for the audio files that belonged to train, validation, and test partitions in the experiments\footnote{\url{https://github.com/Ethio2021/File-names}}. For each experiment, the models were trained five times and the average result was reported.

\begin{table}[t!]
\begin{adjustwidth}{0in}{0in} 
\centering
\caption{Class distribution between the train, validation, and test partitions.}
\begin{tabular}{|c|c|c|c|c|c|c|}\hline 
\bf{Datasets} & \multicolumn{2}{c|}{\bf{Train}} & \multicolumn{2}{c|}{\bf{Test}} & \multicolumn{2}{c|}{\bf{Validation}} \\\hline
Labels   & Positive    & Negative    & Positive       & Negative      & Positive    & Negative   \\\hline
ASED     & 693         & 804         & 199             & 230           & 99         & 115        \\\hline
EMODB    & 95          & 118         & 27             & 34            & 14          & 17         \\\hline
RAVDESS  & 456         & 524         & 140              & 160            & 56         & 64        \\\hline
URDU     & 140         & 138         & 40             & 40            & 20          & 22     \\\hline   
\end{tabular}
\label{distribution}
\end{adjustwidth}
\end{table}

\section{Experiment 1: Comparison of SER Methods for Monolingual SER}  
\subsection{Outline} \label{expt1_outline}
The aim was to carry out an initial comparison of the proposed VGGE model with the two existing models discussed above, AlexNet and ResNet50. Four datasets were used, ASED, RAVDESS, EMO-DB, and URDU.
To allow comparison with the other experiments, the emotion labels for each dataset were mapped onto just two labels, Positive valence and Negative valence, as shown by the scheme in Table \ref{comparison-ASED-RAVDESS-EMO-DB-URDU}. This follows the standard approach found in other work \cite{deng2013sparse, eyben2015geneva}.
When comparing to other papers, we should bear in mind the label mapping which we needed to adopt in order to undertake the later cross-lingual and multilingual experiments. Looking at the table, we can see that the ASED, RAVDESS, EMO-DB and URDU datasets originally had five, eight, seven and four emotion classes respectively, and that these are now being mapped onto just two classes, positive and negative emotion. This simplifies the task, which can account for higher performance figures than in other published works.

Experiment 1 has two parts. In Experiment 1.1, the groups of speakers used for training and testing were varied. In Experiment 1.2, the dataset sentences used for training and testing were varied.

\subsection{Experiment 1.1: Independence of Speakers}

\begin{table}[t]
\begin{adjustwidth}{0in}{0in} 
\centering
\caption{Experiment 1.1: Monolingual SER results for different datasets (train in one language, test in the same language). Two valence values are used for all datasets, created with the mappings shown in Table \ref{comparison-ASED-RAVDESS-EMO-DB-URDU}.} 
\begin{tabular}{|c|c|c|c|c|}
\hline 
\bf{Model} & \bf{ASED}  & \bf{EMO-DB} & \bf{RAVDESS} & \bf{URDU}    \\\hline 
AlexNet  & 78.71 & 68.52 & 80.63 & 93.75 \\ \hline
VGGE     & 84.76 & 85.19 & 83.13 & 70.00 \\ \hline
ResNet50 & 84.13 & 79.63 & 84.38 & 90.00 \\\hline
\textbf{Average} & 82.53 & 77.78 & 82.71  & 84.58 \\
\hline
\end{tabular}
\label{Baseline}
\end{adjustwidth}
\end{table}

\begin{table}[t]
\begin{adjustwidth}{0in}{0in} 
\centering
\caption{Experiment 1.2: Monolingual SER results for the different datasets. Training and testing sentences are varied in this experiment.} 
\begin{tabular}{|c|c|c|c|}
\hline 
\bf{Model} & \bf{ASED}  & \bf{EMO-DB} & \bf{RAVDESS}    \\\hline 
AlexNet  & 80.93 & 55.74 & 82.22 \\ \hline
VGGE     & 86.63 & 70.49 & 83.33 \\ \hline
ResNet50 & 85.82 & 73.77 & 77.78 \\\hline
\textbf{Average} & 84.46 & 66.67 & 81.11 \\
\hline
\end{tabular}
\label{text_ind}
\end{adjustwidth}
\end{table}

Results of the experiment are shown in Table \ref{Baseline}. We can see that VGGE was the best on ASED (Amharic) and EMO-DB (German), ResNet50 was the best on RAVDESS (English) and AlexNet was the best on URDU (Urdu). 

It is interesting to look at the average figures on the bottom row of the table. ASED (82.53\%) and RAVDESS (82.71\%) are very close, EMO-DB (77.78\%) is 4.75\% lower than ASED, and URDU (84.58\%) is 2.05\% higher than ASED. Generally, the differences are not that large when we consider that the languages have very different characteristics, and that the datasets were created independently by different researchers. Moreover, recall that the original data is being mapped onto two sentiment classes from the original four-to-eight classes (see Section \ref{expt1_outline} and Table \ref{comparison-ASED-RAVDESS-EMO-DB-URDU}). 

Subject to these points, we might conclude that Amharic and English monolingual mono-corpus SER are of similar difficulty, German is more difficult and Urdu is easier.
As languages, English and German are perhaps the most similar, since they are both within the Germanic branch of the Indo-European language group. Urdu is also Indo-European, but from the Indo-Iranian branch. Finally, Amharic is from the Semitic branch of the Afro-Asiatic group.
\subsection{Experiment 1.2: Independence of Sentences}
Recall that the datasets all consist of different sentences spoken in every emotion, with the exception of the URDU dataset, based on TV talk show conversation, where individual sentences are not identified. Hence, URDU was not used here.

In this experiment, sentences were either used for training or testing.
For each of the datasets shown in the table, the proposed VGGE model, along with AlexNet and ResNet50, were trained using MFCC features. Each model was trained five times using a 80\%/20\% train/test split, and the average results were computed.

Results are in Table \ref{text_ind}. The trends are similar to Experiment 1.1. This time, VGGE is the best on ASED and RAVDESS, while ResNet50 is the best on EMO-DB. Concerning the averages, ASED and RAVDESS are fairly close (84.46\%, 81.11\%), while EMO-DB is lower (66.67\%). So this again suggests that Amharic and English monolingual SER are of similar difficulty and easier, within the context of these particular datasets and this task, while German SER is more difficult.


\section{Experiment 2: Comparison of SER methods for  Amharic cross-lingual SER}
The aim was to compare the three models AlexNet, VGGE, and ResNet50  (Section \ref{architectures}) when applied to cross-lingual SER. This time, the systems are trained on data in one language and then tested on data in another language. Firstly, the three models are trained on ASED and then tested on EMO-DB, then trained on EMO-DB and tested on ASED, and so on, for different combinations. To allow this cross-training, dataset-specific emotion labels are mapped onto two classes, Positive and Negative, using the same method as for Experiment 1.

Once again, MFCC features were used for all models. The network configuration for VGGE was the same as in the preceding Experiment (Figure \ref{block}). For the other models, the standard configuration and settings were used.
\begin{table}[t]
\begin{adjustwidth}{0in}{0in} 
\centering
        \caption{Experiment 2: Cross-lingual SER results (train in one language, test in another language).}

\begin{tabular}{|c|c|c|c|c|}
 \hline
\bf{Model} & \bf{Training} & \bf{Testing} & \bf{Accuracy}  & \bf{F1-score} \\ \hline
AlexNet & ASED     & EMO-DB  & 65.80    &   56.85       \\
                            & EMO-DB   & ASED    & 62.39    &  58.53  \\
                            & ASED     & RAVDESS & 66.00    & 53.17 \\
                            & RAVDESS  & ASED    & 65.87    & 55.57    \\
                            & ASED     & URDU    & 60.00    & 56.28 \\
                            & URDU     & ASED    & 50.67    & 48.45         \\\hline
                            \textbf{Average}  &    &  &  61.79\%     & 54.81\% \\\hline     
VGGE                        & ASED     & EMO-DB  & 66.67    & 52.55  \\
                            & EMO-DB   & ASED    & 64.22    & 58.53 \\
                            & ASED     & RAVDESS & 59.25    & 51.85     \\
                            & RAVDESS  & ASED    & 61.43    & 62.75 \\
                            & ASED     & URDU    & 59.69    & 56.34\\
                            & URDU     & ASED    & 60.00    & 53.94     \\\hline
                             \textbf{Average}  &    &  &  61.88\%        & 55.99\% \\\hline  
ResNet50                    & ASED     & EMO-DB  & 64.06     &  50.42        \\
                            & EMO-DB   & ASED    & 58.72      & 45.94  \\
                            & ASED     & RAVDESS & 61.75     & 48.68 \\
                            & RAVDESS  & ASED    & 64.16     &   52.66  \\
                            & ASED     & URDU    & 61.56    &   62.06   \\
                            & URDU     & ASED    & 61.33   &  60.03       
\\\hline 
 \textbf{Average}  &    &  &  61.93\%        & 53.30\% \\\hline  
\end{tabular}
\label{Cross-lingual}
\end{adjustwidth}
\end{table}

Results are presented in Table \ref{Cross-lingual}. As line 1 of the table shows, we first trained on ASED and evaluated on EMO-DB (henceforth written ASED$\rightarrow$EMO-DB). VGGE gave the best accuracy (66.67\%), followed closely by AlexNet (65.80\%), and then ResNet50 (64.06\%). For EMO-DB$\rightarrow$ASED, VGGE was best (64.22\%), also followed by AlexNet (62.39\%), and then ResNet50 (58.72\%).

Next, for ASED$\rightarrow$RAVDESS, AlexNet was best (66.00\%), followed by ResNet50 (61.75\%) and VGGE (59.25\%). For RAVDESS$\rightarrow$ASED, AlexNet was best (65.87\%), closely followed by ResNet50 (64.16\%) and then VGGE (61.43\%).

Thirdly, we used ASED$\rightarrow$URDU. Here, ResNet50 was best (61.56\%), followed by AlexNet (60.00\%) and VGGE (59.69\%). For URDU$\rightarrow$ASED, ResNet50 was best (61.33\%), followed by VGGE (60.00\%) and AlexNet (50.67\%).

It is interesting that for ASED$\leftrightarrow$EMO-DB), VGGE was best, for ASED$\leftrightarrow$RAVDESS, AlexNet was best, and for ASED$\leftrightarrow$URDU, ResNet50 was best. What is more, the figures for AlexNet on ASED$\leftrightarrow$RAVDESS in the two directions (66.00\%, 65.87\%, difference 0.13\%) were very close, as were those for ResNet50 on ASED$\leftrightarrow$URDU (61.56\%, 61.33\%, difference 0.23\%), while those for VGGE on ASED$\leftrightarrow$EMO-DB) (66.67\%, 64.22\%, difference 2.45\%) were slightly further apart.

We can therefore conclude that the performance of the three models was very similar overall. This is supported by the average accuracy figures for AlexNet, VGGE and ResNet50 (61.79\%, 61.88\%, 61.93\%) which are also very close, only 0.14\% from the smallest to the biggest.

The results in the table also show the average F1-score performance for VGGE (55.99\%) is higher than that for AlexNet (54.81\%, 1.18\% lower) and ResNet50 (53.30\%, 2.69\% lower). Hence, it is concluded from these results that the prediction performance of VGGE was best, closely followed by AlexNet and then ResNet50. However, the range of F1-scores is small, only 2.69\% from the smallest to the biggest, indicating only a slight difference in performance between different scenarios.


Regarding the results as a whole, two points can be made. First, the accuracy obtained by training on one language and testing on another is surprisingly good. Second, the best language to train on when testing on Amharic seems to vary by model; for AlexNet it is RAVDESS (65.87\%), for VGGE it is EMO-DB (64.22\%) and for ResNet50 it is RAVDESS again (64.16\%).

Finally, we can compare our results for this experiment (Table \ref{Cross-lingual}) with those given for previous cross-lingual studies in Section 2. Generally, they seem comparable. Our average results are around 62\%. In the previous studies we see 56.8\% \cite{albornoz2015emotion}, 57.87\% \cite{latif2018cross}, 65.3\% \cite{latif2019unsupervised}, and 62.5\% \cite{zehra2021cross}. The highest is 71.62\% \cite{xiao2016speech}. In looking at these figures, we must remember that the exact methods and evaluation criteria used in previous experiments vary, so exact comparisons are not possible. Many different languages and datasets are used, emotion labels may need to be combined or transformed in different ways, and so on. Please refer to Section 2 for the details regarding these figures.

\section{Experiment 3: Multilingual SER}
In the previous experiment, we trained in one language and tested in another. In this final experiment, we trained on several non-Amharic languages and then tested on Amharic.

The same three models were used, AlexNet, VGGE and ResNet50, with the same settings and training regime as in the previous experiments.

Table \ref{Multilingual} shows the results. The first three rows for each model show the results when two datasets were used for training, EMO-DB+RAVDESS, EMO-DB+URDU and RAVDESS+URDU. The fourth row uses all three datasets for training, i.e. EMO-DB+RAVDESS+URDU. In all cases, testing is with ASED.

The best overall performance in the table is for VGGE, training with EMO-DB+URDU (69.94\%).  The average figure for VGGE over all the dataset training combinations is also the best (66.44\%).

When RAVDESS is added to EMO-DB+URDU to make EMO-DB+RAVDESS+URDU, the performance of VGGE falls 1.53\% to 68.41\%. In the results presented in Table \ref{Multilingual_1}, the upper right-hand column shows the average accuracy, and the lower right-hand column the average F1-score. In this case, we see that the highest figures over all three models are for all three datasets (67.12\% and 59.79\% respectively).

\begin{table}[t]
\begin{adjustwidth}{0in}{0in} 
\centering
        \caption{Experiment 3: Multilingual SER results (train in two or three non-Amharic languages, test in Amharic).}

\begin{tabular}{|c|c|c|c|c|}
 \hline 
\bf{Model} & \bf{Training} & \bf{Testing} & \bf{Accuracy} & \bf{F1-score} \\\hline
AlexNet & EMO-DB+RAVDESS      & ASED    & 69.06      &61.28     \\
                            & EMO-DB+URDU         & ASED    & 57.23        & 48.38       \\
                            & RAVDESS+URDU        & ASED    & 62.46     & 51.27      \\
                            & EMO-DB+RAVDESS+URDU & ASED    & 69.77      & 62.30          \\\hline 
                             \textbf{Average}  &    &  &  64.63\%         & 55.81\% \\\hline  
VGGE                        & EMO-DB+RAVDESS      & ASED    & 60.50    & 61.12         \\
                            & EMO-DB+URDU         & ASED    & 69.94     &    65.26      \\
                            & RAVDESS+URDU        & ASED    & 66.89    & 64.56         \\
                            & EMO-DB+RAVDESS+URDU & ASED    & 68.41      & 60.17         \\\hline 
                             \textbf{Average}  &    &  &  66.44\%        & 62.78\% \\\hline  
ResNet50                    & EMO-DB+RAVDESS      & ASED    & 61.33       & 43.52          \\
                            & EMO-DB+URDU         & ASED    & 46.24     &  44.57        \\
                            & RAVDESS+URDU        & ASED    & 64.51     & 56.17         \\
                            & EMO-DB+RAVDESS+URDU & ASED    & 63.18         &  62.32       
 \\\hline
  \textbf{Average}  &    &  &  58.82\%         & 51.65\% \\\hline  
\end{tabular}
\label{Multilingual}
\end{adjustwidth}
\end{table}

\begin{table}[t]
\begin{adjustwidth}{0in}{0in} 
\centering
        \caption{Experiment 3: Multilingual SER average results.}

\begin{tabular}{|c|c|c|c|c|c|}
\hline 
\bf{Training} & \bf{Testing} & \bf{AlexNet} & \bf{VGGE} & \bf{ResNet50} & \textbf{Average accuracy}    \\ \hline 
EMO-DB + RAVDESS        & ASED    & 69.06   & 60.5  & 61.33   & 63.63  \\
EMO-DB + URDU           & ASED    & 57.23   & 69.94 & 46.24   &  57.80 \\
RAVDESS + URDU         & ASED    & 62.46   & 66.89 & 64.51   &  64.62 \\
EMO-DB + RAVDESS + URDU & ASED    & 69.77   & 68.41 & 63.18   & 67.12 \\\hline
\hline 
\bf{Training} & \bf{Testing} & \bf{AlexNet} & \bf{VGGE} & \bf{ResNet50} & \textbf{Average F1-score}    \\\hline 
EMO-DB + RAVDESS        & ASED    & 61.00   & 61.21  & 43.44   & 55.31  \\
EMO-DB + URDU           & ASED    & 48.57   & 65.98 & 38.96   &  52.74 \\
RAVDESS + URDU         & ASED    & 51.64   & 64.63 & 56.23   &  57.33 \\
EMO-DB + RAVDESS + URDU & ASED    & 62.45   & 60.28 & 62.99   & 59.79 \\\hline

\end{tabular}
\label{Multilingual_1}
\end{adjustwidth}
\end{table}

However, the interesting result here is that the best accuracy figure in Table \ref{Multilingual} (EMO-DB+URDU$\rightarrow$ASED, VGGE, 69.94\%) is higher than the best accuracy figure in Table \ref{Cross-lingual} with ASED as target, (RAVDESS$\rightarrow$ASED, AlexNet, 65.87\%) by 4.07\%. Moreover, the best overall average accuracy figure in Table \ref{Multilingual} (VGGE, 66.44\%) is higher than the best overall average accuracy figure in Table \ref{Cross-lingual} (ResNet50, 61.93\%) by 4.51\%.

Once again, The results in Table \ref{Multilingual} show the average F1-score performance for VGGE (62.78\%) is higher than that for AlexNet (55.81\%, 6.97\% lower) and ResNet50 (51.65\%, 11.13\% lower). Furthermore, the best overall average F1-score figure in Table \ref{Multilingual} (VGGE, 62.78\%) is higher than the best overall average F1-score figure in Table \ref{Cross-lingual} (VGGE, 55.99\%) by 6.79\%. 

These results suggest that, by using several non-Amharic datasets for training, we can obtain a better result, by several percent, than when using one non-Amharic dataset for training, when testing on Amharic.

Comparing with previous studies in Section 2, there are only three which present multilingual experiments. Lefter et al. \cite{lefter2010emotion} report that training on three datasets, EMO-DB, DES and ENT, and testing on EMO-DB gave the best result, better than their cross-lingual trials. This concurs with our own findings, where average results for Experiment 3 (Table \ref{Multilingual}, bottom line) were higher than those of Experiment 2 (Table \ref{Cross-lingual}, bottom line). Latif et al. \cite{latif2018cross} found that training on EMO-DB, EMOVO and SAVEE and testing on URDU gained a better result than using just two training datasets. Latif et al. \cite{latif2019unsupervised} also obtained the best result when training on three datasets.

\section{Conclusions}
In this paper, we first proposed a variant of the well-known VGG model, which we call VGGE, and then applied AlexNet, VGGE and ResNet50 to the task of Speech Emotion Recognition, focusing on the Amharic language. This was made possible by the existence of the publicly-available Amharic Speech Emotion Dataset (ASED) which we created in previous work \cite{retta2023new}.
In Experiment 1, we trained the three models on four datasets, ASED (Amharic), RAVDESS (English), EMO-DB (German), and URDU (Urdu). In each case, a model was trained on one dataset and then tested on that same dataset. Speaker-independent and sentence-independent training variants were tried. The results suggested that Amharic and English monolingual SER are almost equally difficult on the datasets we used for these languages, German is harder, and Urdu is easier.

In Experiment 2, we trained on SER data in one language and tested on data in another language, for various language pairs. 
When ASED was the target, the best dataset to train on was RAVDESS for AlexNet and ResNet50, and EMO-DB for VGGE. This could indicate that, in terms of SER, Amharic is more similar to English and German than it is to Urdu.


In Experiment 3, we combined datasets for two or three different non-Amharic languages for training, and used the Amharic dataset for testing. The best result in Experiment 3 (EMO-DB+URDU$\rightarrow$ASED, VGGE, 69.94\%) was 4.07\% higher than the best result in Experiment 2 (RAVDESS$\rightarrow$ASED, AlexNet, 65.87\%). In addition, the best overall average figure in Experiment 3 (VGGE, 66.44\%) was 4.51\% higher than the best overall average figure in Experiment 2 (ResNet50, 61.93\%). These findings suggest that if several non-Amharic datasets are used for SER training, the results can be better than if one non-Amharic dataset is used, when testing is on Amharic.
Overall, the experiments demonstrate how cross-lingual and multilingual approaches can be used to create effective SER systems for languages with little or no training data, confirming the findings of previous studies.
Future work could involve improving SER performance when training on non-target languages, and trying to predict which combination of source languages will give the best result.

\section*{Acknowledgments}
This work was supported by the National Key Research and Development Program of China under grant 2020YFC1523300. Many thanks to George Kour for the ArXiv style: https://github.com/kourgeorge/arxiv-style.


%
%
%

\bibliographystyle{abbrv}

\end{document}